\documentclass[conference]{IEEEtran}
\usepackage{comment}
\usepackage{hyperref}
\usepackage{xcolor,soul,framed} %,caption
\newcounter{example}[section]
\newenvironment{example}[1][]{\refstepcounter{example}\par\medskip
   \noindent \textbf{Example~\theexample. #1} \rmfamily}{\medskip}
   
\colorlet{shadecolor}{yellow}
\usepackage{multirow}
\usepackage{amssymb}

\usepackage[cmex10]{amsmath}
\usepackage{mdwmath}
\usepackage{mdwtab}
\usepackage{eqparbox}
\usepackage{url}
\usepackage{smartdiagram}
\usepackage{tikz}
\usetikzlibrary{shapes.geometric, arrows, positioning}
\newcommand{\name}{\textsc{Socrates}}

\usepackage[ruled,vlined,noend]{algorithm2e}

\usepackage{array}
\newcolumntype{H}{>{\setbox0=\hbox\bgroup}c<{\egroup}@{}}

\colorlet{punct}{red}
\definecolor{background}{HTML}{EEEEEE}
\colorlet{delim}{blue}
\colorlet{numb}{red}

\usepackage[T1]{fontenc}
\usepackage{listings}
\usepackage{xcolor}
\lstdefinelanguage{json}{
    basicstyle=\small\normalfont\ttfamily,
    numbers=left,
    numberstyle=\scriptsize,
    stepnumber=1,
    numbersep=8pt,
    showstringspaces=false,
    breaklines=true,
    frame=lines,
    literate=
     *{:}{{{\color{punct}{:}}}}{1}
      {\{}{{{\color{delim}{\{}}}}{1}
      {\}}{{{\color{delim}{\}}}}}{1},
    xleftmargin=0.5cm
}

\begin{document}

\title{\name: Towards a Unified Platform for\\Neural Network Analysis}
\author{Anonymous}  
\author{
Long H. Pham, Jiaying Li and Jun Sun \\
Singapore Management University 
}

\maketitle

% === ABSTRACT ====================================================================
% =================================================================================
\begin{abstract}
Studies show that neural networks, not unlike traditional programs, are subject to bugs, e.g., adversarial samples that cause classification errors and discriminatory instances that demonstrate the lack of fairness. Given that neural networks are increasingly applied in critical applications (e.g., self-driving cars, face recognition systems and personal credit rating systems), it is desirable that systematic methods are developed to analyze (e.g., test or verify) neural networks against desirable properties. Recently, a number of approaches have been developed for analyzing neural networks. These efforts are however scattered (i.e., each approach tackles some restricted classes of neural networks against certain particular properties), incomparable (i.e., each approach has its own assumptions and input format) and thus hard to apply, reuse or extend. In this project, we aim to build a unified framework for developing techniques to analyze neural networks. Towards this goal, we develop a platform called \name~which supports a standardized format for a variety of neural network models, an assertion language for property specification
as well as multiple neural network analysis algorithms including two novel ones for falsifying and probabilistic verification of neural network models. \name~is extensible and thus existing approaches can be easily integrated. Experiment results show that our platform can handle a wide range of networks models and properties. More importantly, it provides a platform for synergistic research on neural network analysis. 
\end{abstract}

% === KEYWORDS ====================================================================
% =================================================================================
% \begin{IEEEkeywords}

% \end{IEEEkeywords}

\section{Introduction}\label{intro}
Neural network models are getting ever more popular due to their exceptional performance in solving many real-world problems, such as self-driving cars~\cite{driving2016cars}, face recognition~\cite{yin2017multi}, malware detection~\cite{malware2014dtc}, sentiment analysis~\cite{tang2015document} and machine translation~\cite{machine2014translation}. At the same time, neural networks are shown to be vulnerable to a variety of issues. For instance, it is shown that adversarial perturbation can be applied to generate samples which trigger wrong model prediction~\cite{szegedy2013intriguing,goodfellow2014explaining,cw2017Robustness}; and it is shown that neural network models may discriminate certain groups or individuals~\cite{zhang2020white}. Given that neural networks are increasingly applied in applications which are safety-critical (e.g., self-driving cars) or have significant societal impact (e.g., personal credit rating systems or face recognition), it is desirable that such neural network models are systematically analyzed against a variety of desirable properties.
% For instance, an image recognition neural network model used in a self-driving car should be verified to be robust (i.e., the classification result remains the same in the presence of perturbation) and a neural network for predicting personal credit rating should be verified to be fair. 

Recently, there has been an increasing number of efforts on formally analyzing neural network models. In~\cite{katz2017reluplex}, Katz \emph{et al.} proposed a constraint solving technique targeting feedforward neural networks with ReLU activation functions. In~\cite{wang2018formal,DBLP:conf/nips/WangPWYJ18}, Wang \emph{et al.} improved the constraint solving techniques for verifying the same class of neural networks with symbolic intervals. In~\cite{gehr2018ai2,singh2018fast,singh2019abstract}, the authors applied the abstract interpretation techniques to verify neural networks with activation functions such as ReLU, Sigmoid and Tanh. Besides that, there are work focus on testing and attacking neuron networks~\cite{deepXplore,cw2017Robustness}.

The status quo is however less than satisfactory.
% , i.e., existing approaches are limited in multiple ways, which makes applying, comparing, reusing and extending existing verification efforts difficult.
First, each existing approach supports only restricted classes of neural networks or properties. For instance, some existing work~\cite{katz2017reluplex,DBLP:conf/nips/WangPWYJ18} only support verification of
% neural network models other than 
feedforward neural networks using ReLU activation functions. Only very recently, researchers have started exploring the verification of feedforward neural networks with different activation functions~\cite{singh2018fast,singh2019abstract} and some subclasses of recurrent neural networks~\cite{DBLP:journals/corr/abs-2004-02462,DBLP:journals/corr/abs-2005-13300}. Furthermore, existing verification approaches normally focus on reachability properties or local robustness, and ignores other important properties such as fairness and beyond. Secondly, existing toolkits require input models in specific format and different tools often require different format. For instance, Reluplex~\cite{katz2017reluplex} requires a text file contains the weights and biases of multilayer perceptron layers, whereas DeepPoly~\cite{singh2019abstract} needs a more complex input which specifies the types of layers before providing their parameters' values. This will likely get worse as there are an increasing list of popular frameworks for training neural network models, such as TensorFlow, Cafe, MXNet, PyTorch, Theano and Keras, all of which encode neural network models in their own ways. As a result, a tool developed for one framework may not be applicable to models trained using another.
% This not only limits the applicability of the existing verification toolkits but also makes comparing them infeasible.
We remark that two recent efforts on solving this problem are ONNX~\cite{onnx} and NNEF~\cite{nnef}, which aims to provide a cross-platform format for neural networks. It is however designed for a different purpose and lacks important features which are required for neural network analysis. Thirdly, each existing approach typically focuses on one property, whereas in fact the algorithm could be easily extended to analyze neural networks against another property. For instance, algorithms for testing or verifying local robustness can easily be extended to test or verify fairness defined in terms of individual discrimination, which has been demonstrated in~\cite{zhang2020white}. With different verification algorithms and testing techniques implemented in different repositories based on different input format, reusing existing efforts is often nontrivial. 

In this project, we aim to build a unified framework for developing analysis techniques for neural networks. The goal is to have a platform which allows us to apply, compare, reuse and develop techniques for a variety of neural network models against a variety of properties. Towards this goal, we design and implement an open source platform called \name, which embodies multiple technical contributions. First, \name~provides a standardized format for a variety of neural network models based on JSON. By compiling models trained using different frameworks to this common format and building analysis algorithms around it, the same algorithm can be applied to models trained using different frameworks. Secondly, \name~supports an assertion language which is designed to specify a range of properties of neural network models, including robustness, fairness and more. Thirdly,  \name~provides multiple analysis algorithms, including two novel ones, 
i.e., optimization-based falsification and statistical model checking, which can be applied to verify or falsify a variety of neural network models. More importantly, \name~is designed to be modular and extensive, i.e., it is easy to support new models, properties or algorithms.
% ; or integrate existing verification engines.  

Furthermore, we provide a comprehensive set of 12347 tasks (i.e., neural network models and respective assertions) as a part of the \name~repository so that researchers can easily evaluate and compare the effectiveness and efficiency of different analysis algorithms using a comprehensive set of benchmarks. As an example, we use the benchmarks and conduct multiple experiments to evaluate the effectiveness of the two new algorithms that we developed in \name. The experiment results show that the new algorithms can handle a wide range of networks models and properties. We remark that \name~is open source at~\cite{srcurl} and we are making all the effort required to develop it into a platform for synergistic research on analysis of neural networks. 

The rest of the paper is organized as follows. In Section~\ref{sec:overall}, we discuss the overall design of \name. In Section~\ref{sec:json}, we present details on the model format in JSON supported in \name. In Section~\ref{sec:algos}, we present the two new algorithms supported in \name~and evaluate their effectiveness against our benchmark. In Section~\ref{sec:related}, we review related work and we conclude in Section~\ref{sec:conclude}.
\section{System Overview}
\label{sec:overall}
\name~is designed for both ordinary users who require a tool for analyzing a particular neural network model as well as researchers who are working on developing neural network analysis techniques. In the following, we first illustrate how \name~works from an ordinary user point of view and then introduce its design from a researcher point of view. Lastly, we provide an overview of functionalities provided by \name.%~and those by existing toolkits for neural networks.  

\subsection{For Ordinary Users} \label{caseStudy}

To use \name~to analyze a neural network model against certain property, a user must provide a JSON file which encodes the analysis task in the required format. A task is composed of three main parts, i.e., a model, a property, and an engine selected for solving the task. The JSON format is designed to support a variety of neural network models.
% (see Section~\ref{model} for details on the supported network models).
The model part of a JSON file can be generated automatically from models trained using existing frameworks such as Tensorflow and PyTorch. The property part of the JSON file is specified in an assertion language designed for neural network analysis, with a formal syntax as well as supporting easy-to-use templates for commonly used properties.
Together with the model and the property are JSON keys for specifying the analysis engine. Note that some engines have configurable parameters, which are specified as part of the JSON file as well. The readers are referred to Section~\ref{model} and~\ref{assertion} for details on the format. Once the JSON file is loaded, the user simply waits for the result.   

In the following, we use an example to illustrate the process. Assume that the user has trained a network for classifying images with 1 channel and a dimension of $28 \times 28$ (i.e., 784 pixels in total) based on the MNIST dataset. The design of the neural network is shown in Figure~\ref{enet}. It is a multilayer perceptron with 4 layers. Beside the input layer, each of the two hidden layers has 50 neurons, and the output layer has 10 neurons. The activation function used in two hidden layers is the ReLU function, and the output layer uses the softmax function to return the probability of each class according to the input sample. The property to be analyzed is local robustness, i.e., given a particular input image, with a limit on the value of each pixel to be perturbed, all perturbed images have the same label as the original one. 

\tikzset{%
  every neuron/.style={
    circle,
    draw,
    minimum size=1cm
  },
  neuron missing/.style={
    draw=none, 
    scale=4,
    text height=0.333cm,
    execute at begin node=\color{black}$\vdots$
  },
}

\begin{figure}[t]
\centering
\begin{tikzpicture}[x=1.5cm, y=1.5cm, >=stealth, scale=0.6, every node/.style={scale=0.6}]

\foreach \m/\l [count=\y] in {1,2,3,missing,4}
  \node [every neuron/.try, neuron \m/.try] (input-\m) at (0,2.5-\y) {};

\foreach \m [count=\y] in {1,missing,2}
  \node [every neuron/.try, neuron \m/.try ] (hidden1-\m) at (2,2-\y*1.25) {};
  
\foreach \m [count=\y] in {1,missing,2}
  \node [every neuron/.try, neuron \m/.try ] (hidden2-\m) at (4,2-\y*1.25) {};

\foreach \m [count=\y] in {1,missing,2}
  \node [every neuron/.try, neuron \m/.try ] (output-\m) at (6,1.5-\y) {};

\foreach \l [count=\i] in {1,2,3,784}
  \draw [<-] (input-\i) -- ++(-1,0)
    node [above, midway] {$I_{\l}$};

\foreach \l in {1}
  \node [above] at (hidden1-1.north) {$H1_{\l}$};

\foreach \l in {50}
  \node [below] at (hidden1-2.south) {$H1_{\l}$};
  
\foreach \l in {1}
  \node [above] at (hidden2-1.north) {$H2_{\l}$};

\foreach \l in {50}
  \node [below] at (hidden2-2.south) {$H2_{\l}$};
  
\foreach \l [count=\i] in {1,10}
  \draw [->] (output-\i) -- ++(1,0)
    node [above, midway] {$O_{\l}$};

\foreach \i in {1,...,4}
  \foreach \j in {1,...,2}
    \draw [->] (input-\i) -- (hidden1-\j);

\foreach \i in {1,...,2}
  \foreach \j in {1,...,2}
    \draw [->] (hidden1-\i) -- (hidden2-\j);

\foreach \i in {1,...,2}
  \foreach \j in {1,...,2}
    \draw [->] (hidden2-\i) -- (output-\j);

\foreach \l [count=\x from 0] in {Input, 1st Hidden, 2nd Hidden, Output}
  \node [align=center, above] at (\x*2,2) {\l \\ layer};

\end{tikzpicture}
\caption{An example neuron network}
\vspace{-4mm}
\label{enet}
\end{figure}
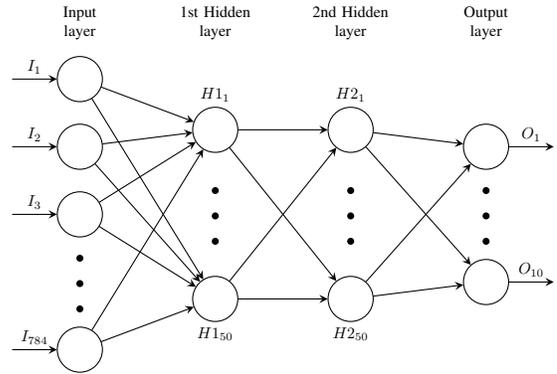

The JSON file is shown in Figure~\ref{jsoninput}. 
At line 2, a model is defined with the key \emph{model}. The value of \emph{model} is a JSON object. At line 3, the value of the key \emph{shape} is specified as an integer tuple which represents the shape of the input sample. For this example, the value is $(1, 784)$ where $1$ is the length of the sample and $784$ is the size of each element in the sample.
At line 4, the key \emph{bounds} specifies the lower bound and the upper bound respectively for features in valid input samples.
In this example, the bound $[(0,1)]$ means each of the features in a valid input (i.e., a feature vector) is a value $v$ which satisfies $0 \leq v \leq 1$. Lines 5 to 24 then specify the value of the key \emph{layers}, i.e., an array in which each element specifies the details of one layer. In this example, the array contains information of the two hidden layers and the output layer. Note that the details of the input layer are not necessary (since they have been specified using other keys). Each array element specifies the type of the layer, the value of the weight matrix, the value of the bias vector and the activation function. These information are defined using the keys \emph{type}, \emph{weights}, \emph{biases} and \emph{func} respectively. We remark that the values of \emph{weights} and \emph{biases} are omitted as they are rather complicated, e.g., the value of \emph{weights} for the first hidden layer is a matrix of dimension $50 \times 784$ written as a string with a syntax similar to the right-hand side of Python multidimensional array initialization. 
Instead of providing these values directly, the user can provide addresses to the files (either local or online) containing their values.
Note that the key values might be correlated and thus must be checked for well-formness, i.e., the second dimension of the weight matrix must be equal to the number of neurons in the previous layer; and the first dimension
of the weight matrix and
the size of the bias vector
must be equal to the number of neurons in the current layer.

\begin{figure}[t]
\centering
\includegraphics[scale=0.15]{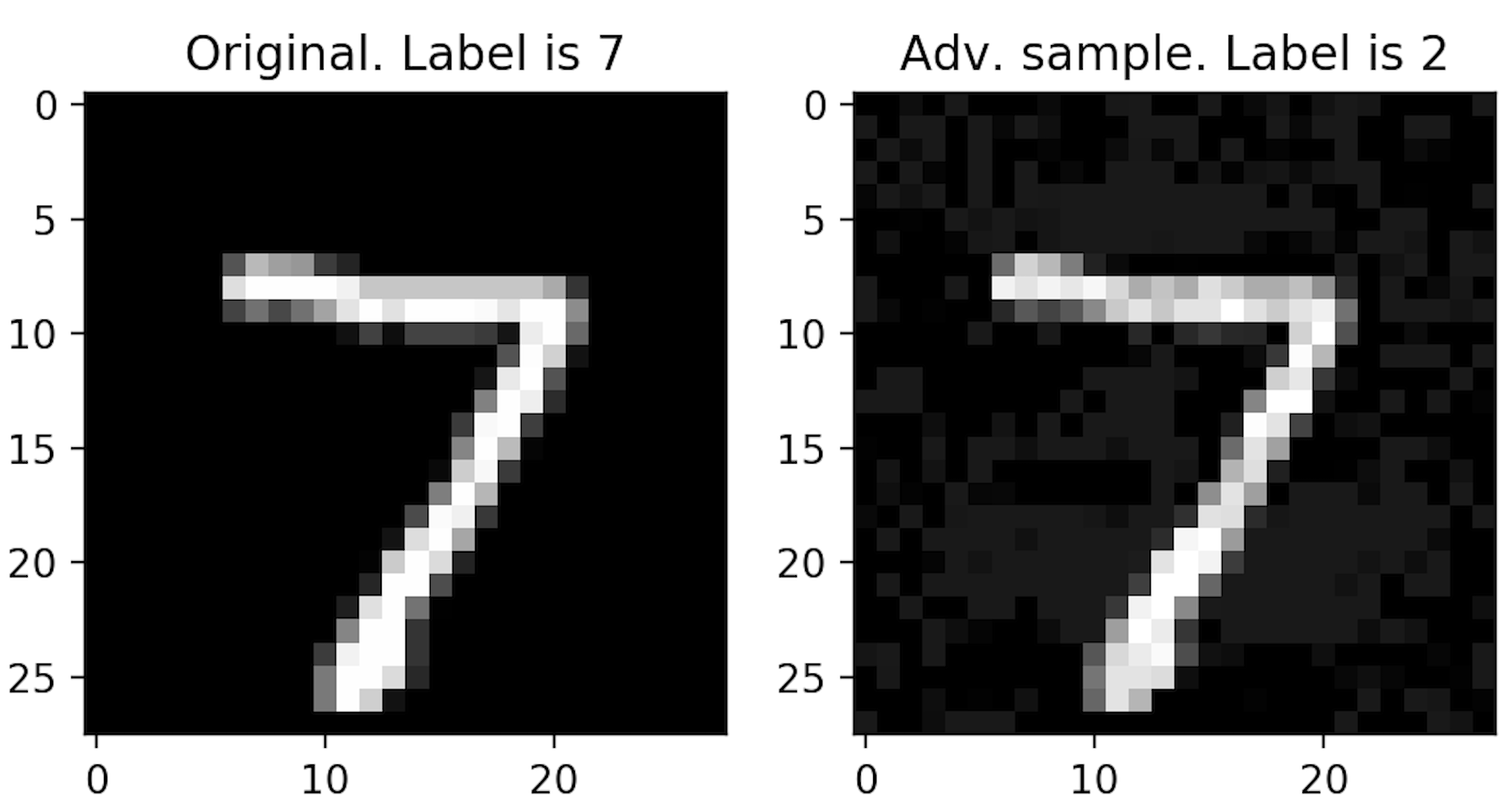}
\caption{Example analysis results}
\vspace{-2mm}
\label{mov}
\end{figure}

Lines 26 to 31 then define the property to be analyzed. In this example, the property is local robustness which is specified with multiple keys. The key \emph{x0} is an input image, i.e., a vector of size 784, or an address to an image.
% (either local or online). %That is, local robustness is defined based on an input. 
In this example, assume that the input image is the one shown on the left of Figure~\ref{mov}, i.e., an image from the MNIST dataset with label 7.
The key \emph{distance} specifies the searching space using one of the predefined functions in the framework. In this example, the distance function is the infinity norm distance, which intuitively means the maximum element-wise absolute difference between two samples. Finally, the key \emph{eps} specifies the maximum value of the distance. In this example, a predefined template for local robustness is used. \name~supports a general assertion language, which we present in Section~\ref{assertion}. 

\begin{figure}[t]
\begin{lstlisting}[language=json,firstnumber=1,basicstyle=\footnotesize]
{
  "model": {
    "shape": "(1,784)",
    "bounds" : "[(0,1)]",
    "layers": [
      {
        "type": "linear",
        "weights": "a [50 x 784] matrix",
        "biases": "a [50] vector",
        "func": "ReLU"
      },
      {
        "type": "linear",
        "weights": "a [50 x 50] matrix",
        "biases": "a [50] vector",
        "func": "ReLU"
      },
      {
        "type": "linear",
        "weights": "a [10 x 50] matrix",
        "biases": "a [10] vector",
        "func": "softmax"
      }
    ]
  },
  "assert": { 
    "robustness": "local",
    "x0": "a [784] vector",
    "distance": "infinite norm",
    "eps": "0.1"
  },
  "solver": {
    "algorithm": "optimize"
  },
  "display": {
    "resolution": "(28,28)"
  }
}
\end{lstlisting}
\caption{An example JSON input}
% \vspace{-2mm}
\label{jsoninput}
\end{figure}

Lines 32 to 34 then specify the engine that is applied to solving the problem. In this example, the user selects an optimization-based falsifier. The falsifier automatically transforms the problem into an optimization problem, which is then solved using an optimization algorithm (see details in Section~\ref{solver}). An image which violates the specified property is identified in about 1 second, which is shown on the right of Figure~\ref{mov}. That is, this image has a distance from the original image not greater than the specified bound (i.e., $0.1$ in this example) and has a label which is not 7. Note that the JSON file also contains keys (at lines 35-37) which are for presenting the falsification result, e.g., by setting the \emph{display} function and providing a \emph{resolution}.

\subsection{For Researchers} \label{architecture}
More relevantly, \name~is designed to enable synergistic effort on developing state-of-the-art analysis techniques for neural networks. It has a publisher-subscriber architecture which facilitates developing analysis techniques for different neural network models and properties independently. It has three main modules, i.e., the \emph{Parser} and \emph{Display} on the front-end and \emph{Analysis Engines} on the back-end. In the following, we briefly discuss the main functionality of each component. The technical details are discussed in the subsequent sections.

The \emph{Parser} module is built upon the JSON format (as exemplified in Figure~\ref{jsoninput}). The parser receives a JSON file, checks its well-formness, decodes it and publishes it internally as a task.
Note that each task is associated with a number of parameters (such as the network model type, size and the assertion type), which are used to determine whether an engine is applicable or not. The models which are currently supported in \name~include multilayer perceptron (MLP), convolutional networks (CNN), residual networks (ResNet), and recurrent networks (RNN). To support new models, the developers are required to extend the JSON format (by introducing new values for existing keys or introducing new keys) and extend the parser accordingly to generate the task.%, which is relatively easy.     

The \emph{Analysis Engines} module consists of a set of engines which could be developed independently from each other. Each engine could be either general or specific (i.e., dedicated to certain models or certain properties, such as many existing verification engines like DeepPoly~\cite{singh2018fast,singh2019abstract}). For instance, the two new engines supported in \name~applies to all neural network models which are currently supported in \name. The first one is an optimization-based falsification engine, which transforms the task into the problem of finding a counterexample through optimization. The second one is a statistical model checking~\cite{DBLP:conf/atva/ClarkeZ11} engine which can be used to verify that the assertion holds with certain level of statistical confidence. As the engines are independent, extending \name~with a new algorithm is straightforward.
% As an example, it only took us two days to integrate DeepPoly, a verifier for local robustness property for feedforward neural networks, into \name.    

The \emph{Display} module is used to present the results in a user-friendly way to the user. An analysis engine typically generates three kinds of results, i.e., the property is verified, no counterexample is identified (e.g., which timeout occurs) or a counterexample is identified. Depending on the application domain and the property, the counterexample could be an image, a text, a feature vector or a set of them (for instance, if the property to be analyzed is individual fairness, two contrasting feature vectors form a counterexample). The display module receives the result from the engines and displays them accordingly. Note that some additional keys are defined in the JSON format so that the user can specify the display options. 

\begin{figure}[t]
\centering
\includegraphics[scale=0.08]{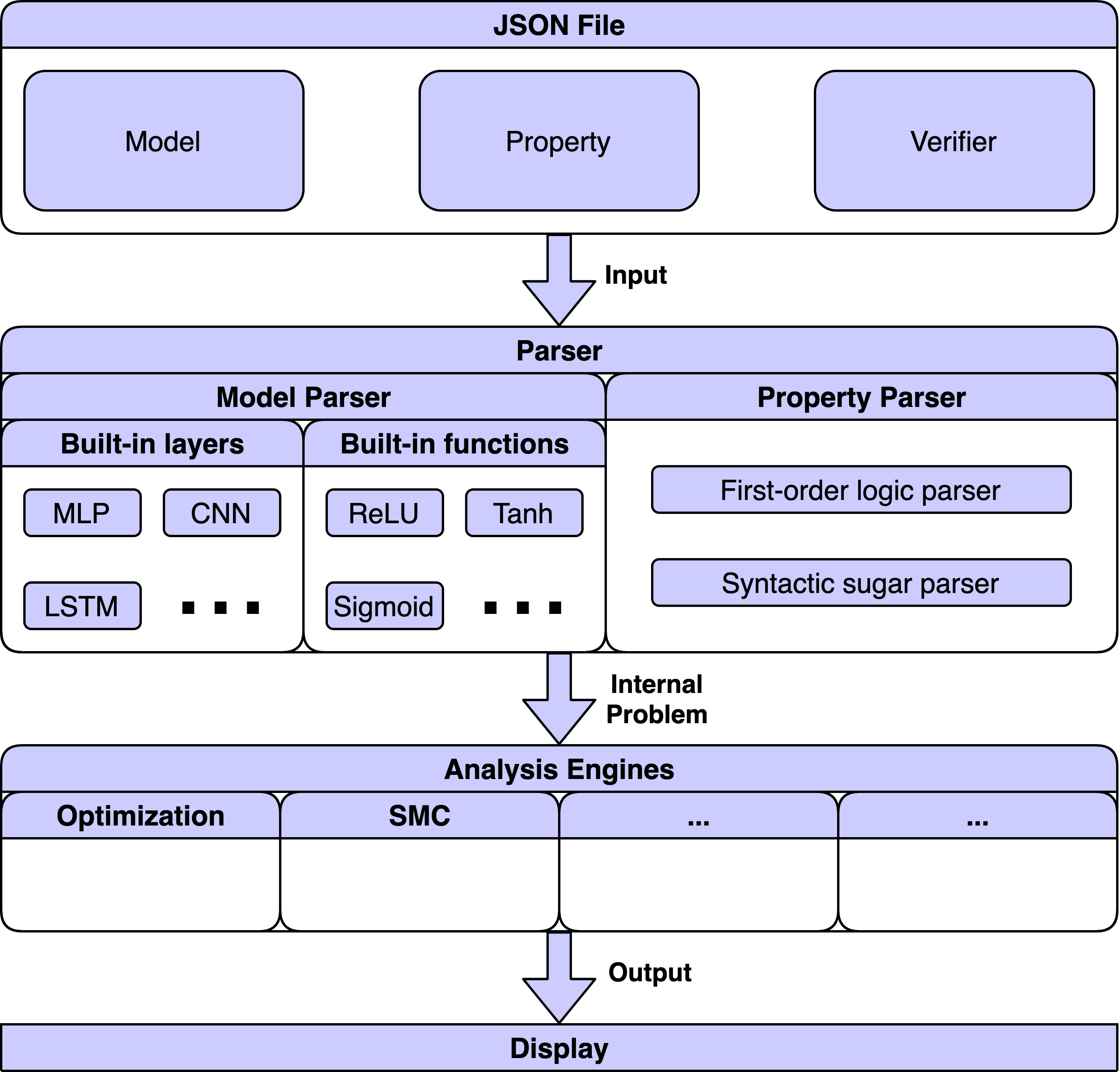}
\caption{The architecture of \name}
\vspace{-2mm}
\label{arch}
\end{figure}

\subsection{Functionalities}
\label{func}
\name~is designed to be a unified platform supporting a variety of neural network models, properties and algorithms. 
In the following, we compare \name~with existing state-of-the-art approaches in terms of functionalities. There are recently a booming number engines for neural networks and thus it is hard to keep up with all of them. Our search of state-of-the-art tools are based on research papers recently published at top-tier conferences and it is possible that we might miss some of them. Furthermore, not all tools reported in the publications are available for evaluation (or reliable enough to be evaluated independently). The following are the tools that we gather and compare: Reluplex~\cite{katz2017reluplex}, Neurify~\cite{DBLP:conf/nips/WangPWYJ18}, DeepZ~\cite{singh2018fast}, DeepPoly~\cite{singh2019abstract}, RefineZono~\cite{singh2018boosting}, RefinePoly~\cite{singh2019beyond},
ADF~\cite{zhang2020white}, and C\&W~\cite{cw2017Robustness}. Note that the last two are testing engines while the others are verification engines. 
They are included as representatives of their kind to provide benchmark on the state-of-the-art for verification or testing.
Their results however should be taken with a grain of salt as they are designed for completely different purpose. 
To find out the functionalities of each tool, we check the papers which reported the tool and experiment the tool to find out whether new capabilities have been introduced recently. For each tool, we identify the kind of supported neural network models and properties. 

The comparison is summarized in Table~\ref{nets}, where $\blacktriangle$ means partial support (i.e., only works with small networks and may not be scallable to bigger ones). In terms of models supported by the tools, it can be observed that \name~is the only tool which supports all types of neural networks. For Reluplex, experiments on verifying small multilayer perceptron networks have been reported. For Neurify, besides multilayer perceptron, it also supports convolutional networks. The next four tools, i.e., DeepZ, DeepPoly, RefineZono and RefinePoly, have the same capability, i.e., they support multilayer perceptron, convolutional, and residual networks.
% For DeepPoly, it does support recurrent networks (other than unfolding a recurrent network model multiple times to become a feed-forward neural network) or residual networks. DeepZ supports feed-forward neural network as well as some residual networks.
For ADF, the reported experiments only deal with small multilayer perceptron networks.
% , which is similar to Reluplex. 
All of the adversarial sample generation tools that C\&W represents focus on the image recognition problem, and thus do not support recurrent networks.

\begin{table*}[t]
\small
\begin{center}
\begin{tabular}{| l | c | c | c | c | c | c | c | c |}
\hline
 \multirow{2}{*}{Tools} & \multicolumn{4}{c|}{Types of networks} & \multicolumn{4}{c|}{Types of properties} \\
\cline{2-9}
  & MLP & CNN & ResNet & RNN & Local robust. & Global robust. & Lin. (in)equal. & Fairness \\
\hline
 \name & $\bigstar$ & $\bigstar$ & $\bigstar$ & $\bigstar$ & $\bigstar$ & $\bigstar$ & $\bigstar$ & $\bigstar$ \\
 Reluplex & $\blacktriangle$ & $\bigcirc$ & $\bigcirc$ & $\bigcirc$ & $\bigstar$ & $\blacktriangle$ & $\bigstar$ & $\bigcirc$ \\
 Neurify & $\bigstar$ & $\bigstar$ & $\bigcirc$ & $\bigcirc$ & $\bigstar$ & $\bigcirc$ & $\bigstar$ & $\bigcirc$ \\
 DeepZ & $\bigstar$ & $\bigstar$ & $\bigstar$ & $\bigcirc$ & $\bigstar$ & $\bigcirc$ & $\bigstar$ & $\bigcirc$ \\
 DeepPoly & $\bigstar$ & $\bigstar$ & $\bigstar$ & $\bigcirc$ & $\bigstar$ & $\bigcirc$ & $\bigstar$ & $\bigcirc$ \\
 RefineZono & $\bigstar$ & $\bigstar$ & $\bigstar$ & $\bigcirc$ & $\bigstar$ & $\bigcirc$ & $\bigstar$ & $\bigcirc$ \\
 RefinePoly & $\bigstar$ & $\bigstar$ & $\bigstar$ & $\bigcirc$ & $\bigstar$ & $\bigcirc$ & $\bigstar$ & $\bigcirc$ \\
 ADF & $\blacktriangle$ & $\bigcirc$ & $\bigcirc$ & $\bigcirc$ & $\bigcirc$ & $\bigcirc$ & $\bigcirc$ & $\bigstar$ \\
 C\&W & $\bigstar$ & $\bigstar$ & $\bigstar$ & $\bigcirc$ & $\bigstar$ & $\bigcirc$ & $\bigcirc$ & $\bigcirc$ \\
\hline
\end{tabular}
\caption{Comparison of supported networks and properties. $\bigstar$: Full support, $\blacktriangle$: Partial support, $\bigcirc$: No support}
\vspace{-4mm}
\label{nets}
\end{center}
\end{table*}

The properties that each tool can handle are shown in the last column in Table~\ref{nets}. We again observe that only \name~supports all types of properties. Most of the other tools support the local robustness property, except ADF. Reluplex reportedly supports verification of the global robustness property for small networks while other tools do not support global robustness. Most of the tools can support reachability properties expressed using linear (in)equalities, except ADF and C\&W. Besides ADF which supports falsification of fairness properties (through a combination of clustering and searching), fairness properties are only supported by \name.

\subsection{Implementation} \label{impAndEva}

% \subsection{Implementation} \label{implementation}
% In this subsection, we present relevant implementation details which are necessary to extend nSolver with new models, solving techniques or optimization techniques. 
\name~is open source at~\cite{srcurl}, including all the source code as well as a set of 12347 analysis tasks. \name~is implemented using Python 3 with a total of 2400 lines of code. Multiple public Python libraries are used in the implementation, including \emph{json} and \emph{ast} for parsing the JSON input file, \emph{numpy} and \emph{autograd.numpy} for mathematical computation, \emph{matplotlib} for displaying imagery results, and \emph{sicpy} for solving the optimization problem.
% presented in Section~\ref{solver}. 
% By default, the local optimization function (instead of the global one) in \emph{scipy} is applied, although this can be easily configured. %/ Moreover, by default, if the optimization problem only contains bound constraints, we use L-BFGS-B method. For the optimization problem with linear (in)equalities and fairness constraints, we use SLSQP method.
% \sj{The part on implementation is a bit too simple. Do we have more details that we can discuss. For instance, any details related to making it extensible?}

\name~is designed with extensiblity in mind. The types of network models supported by \name~can be easily extended by introducing new types of layers and activation functions. Currently, each type of layer is implemented as an independent class. In each class, the most important function is $apply$, which compute the output vector from the input vector. To extend the capability of \name~with new type of layer, a new class can simply added into the library. Similarly, all the supported activation functions are kept inside an utility class. The new functions can be easily added at any time. The new types of layers and activation functions may need new parameters, which should not be the problem because users can always define the new keys for the JSON input file and update the JSON parser to represent the necessary parameters.

% For the property, as presented in previous sections, property in
In \name, the property can be a first-order logic formula composed from a set of predefined functions or a map contains syntactic sugar definitions (see Section~\ref{assertion} for more details). \name~has a parser (independent from the above JSON parser) implemented with ANTLR to parse general first-order logic formula in the form of string, and then returns an AST to represent the formula. The expressiveness of the formula can be extended by adding new predefined functions and update the parser. Otherwise, users can define new syntactic sugar keys to model new properties.

Finally, similar to the different neural network layers, each engine in \name~is an independent class. So new analysis engines can be easily added as new classes. The most important function in these classes is $solve$, which receives a model and a property, and then applies a specific analysis algorithm. Each engine can be designed to solve general problem with property in form of first-order logic formula or just its own specific problem defined with its own set of syntactic sugar keys. Moreover, each engine may have its own configurable parameters. As explained previously, these parameters are easily defined using new keys and then can be provided in JSON input file.

\section{The JSON Inputs}
\label{sec:json}

% In this section, we introduce the details of the JSON format designed for neural network verification.

Existing efforts on neural network analyzing techniques have resulted in multiple impressive tools, including verifiers like DeepPoly~\cite{singh2019abstract} and Reluplex~\cite{katz2017reluplex}, or testing tools like C\&W~\cite{cw2017Robustness}. These tools, however, have their own input format. For instance, Reluplex only supports multilayer perceptron networks and requires a text file containing the number of layers, the number of neurons in each layer, the normalization information, as well as weights and biases of each layer in the network. DeepPoly, on the other hand, supports different types of networks and thus it requires the type of each layer before the values of the layer's parameters (e.g., weights and biases). Moreover, because DeepPoly focuses on local robustness, it allows users to define the distance value between samples explicitly, whereas Reluplex does not have this feature. 
As a result, it is hard to compare the performance of different tools or to combine techniques developed by different research groups. Furthermore, existing toolkits are often developed for specific properties. For instance, DeepPoly is designed for verifying local robustness; Reluplex focuses on reachability properties; and ADF focuses on falsifying a particular notion of fairness only. A close investigation however shows that an algorithm developed in one tool (e.g., the one developed in ADF) could be potentially extended to other properties (e.g., local or global robustness). Thus, we develop a JSON format which supports a variety of neural network models and an assertion language, to provide a common ground for different neural network analysis techniques. 

\subsection{Specifying Models} \label{model}
The JSON file is composed a sequence of keys which specifies the details of the network model and the property. The keys used to define the model are shown in Table~\ref{keys}. 
At the top-level, a model is specified using the key \emph{model}. The value of \emph{model} is then defined as a JSON object using a triple consisting of keys \emph{shape}, \emph{bounds}, and \emph{layers}. The key \emph{shape} is used to define the shape of the input sample, which is a tuple containing multiple integer values. The first value represents the length of the sample. Note that the first value is always $1$ for non-recurrent neuron networks and is $n$ with $n \geq 1$ for recurrent neural networks. The remaining numbers represent the shape for each element in the sample. 
For instance, the tuple $(5, 80)$ means that the input sample is a sequence with length $5$ and each element in the sequence is a vector with size $80$.

\begin{table}[t]
\begin{center}
\small
\begin{tabular}{| l | l |}
\hline
 Key & Definition \\ 
 \hline
\emph{model} & The details of the network model \\
\emph{shape} & The shape of input sample \\
\emph{bounds} & The bounds of input sample \\
\emph{layers} & The details of model layers \\
\emph{type} & The type of a layer \\
\emph{weights} & The weight matrix of a layer \\
\emph{biases} & The bias vector of a layer \\
\emph{func} & The activation/transformation function of a layer \\
\emph{filters} & The filter matrix of a convolutional layer \\
\emph{padding} & The padding value of a convolutional layer \\
\emph{stride} & The stride value of a convolutional layer \\
\emph{h0}$/$\emph{c0} & The \emph{h0}$/$\emph{c0} vector of a recurrent layer \\
% \emph{c0} & The $c0$ vector of a recurrent layer \\
% \emph{path} & The path to the pre-trained model \\
 \hline
\end{tabular}
\caption{The keys used to define the network model}
\vspace{-4mm}
\label{keys}
\end{center}
\end{table}

\begin{table*}[t]
\small
\begin{center}
\begin{tabular}{| l | l | l |}
\hline
 Type & Definition & Parameters \\ 
 \hline
Linear & A fully-connected layer & 1 weight matrix, 1 bias vector, 1 optional activation function name \\
MaxPool1d & A 1-dimensional max pooling layer & 1 stride value, 1 padding value \\
MaxPool2d & A 2-dimensional max pooling layer & 1 stride value, 1 padding value \\
MaxPool3d & A 3-dimensional max pooling layer & 1 stride value, 1 padding value \\
Conv1d & A 1-dimensional convolutional layer & 1 filter matrix, 1 bias vector, 1 stride value, 1 padding value \\
Conv2d & A 2-dimensional convolutional layer & 1 filter matrix, 1 bias vector, 1 stride value, 1 padding value \\
Conv3d & A 3-dimensional convolutional layer & 1 filter matrix, 1 bias vector, 1 stride value, 1 padding value \\
ResNet2l & A 2-layers residual block & 2(+1) filter matrices, 2(+1) bias vectors, 2(+1) stride values, 2(+1) padding values \\
ResNet3l & A 3-layers residual block & 3(+1) filter matrices, 3(+1) bias vectors, 3(+1) stride values, 3(+1) padding values \\
RNN & A basic RNN layer & 1 weight matrix, 1 bias vector, 1 $h_0$ vector, 1 optional activation function \\
LSTM & A LSTM layer & 1 weight matrix, 1 bias vector, 1 $h_0$ vector, 1 $c_0$ vector \\
GRU & A GRU layer & 2 weight matrices, 2 bias vectors, 1 $h_0$ vector, 1 $c_0$ vector \\
Function & A standalone function layer & 1 function name, other optional parameters for the function \\
 \hline
\end{tabular}
\caption{The different types of layers supported by \name}
\vspace{-4mm}
\label{types}
\end{center}
\end{table*}

The key \emph{bounds} is used to define the bounds of values for valid input samples. Its value is a string representing an array containing tuples. Suppose the number of features in the input sample is $n$ and the number of tuples in the bounds array is $m$, we require that $(n \mod m) = 0$. With the above condition, the $i$-th element in the bounds array is used to constraint the values of input features from index $\frac{i * n}{m}$ to index $\frac{(i + 1) * n}{m} - 1$. Each element in the array then is a tuple containing two numbers $a_i$ and $b_i$ such that $a_i \leq b_i$. The first number $a_i$ is the lower bound and the second number $b_i$ is the upper bound of the corresponding input features. 

The key \emph{layers} specifies details of the network layer-by-layer. Its value is an array. Each element in the array is a JSON object which represents a layer in the network with a specific \emph{type}. According to each type, a layer may contain multiple keys to specify parameters of the layer. In general, each layer can be considered as a function.
% , which produces an output from an input. 
The original sample of the network is the input of the first layer, the output of layer $n$ is the input of layer $n + 1$, and the output of the last layer is the output of the whole network.
% Each layer can be a variety of types.
Table~\ref{types} summarizes the types of layers that are currently supported in \name~as well as their definitions and parameters.%\footnote{Each ResNet2l and ResNet3l layer may have 1 optional filter matrix, bias vector, stride value, and padding value.}.

Each type is associated with a list of parameter values which are specified with one or more predefined keys. For instance, keys \emph{weights}, \emph{biases}, and \emph{func} are used to define the weight matrix, the bias vector, and the activation function respectively.
% as shown in Figure~\ref{jsoninput}.
Depending on the type of the layer, keys such as \emph{filters}, \emph{stride}, \emph{padding}, \emph{h0} and \emph{c0} are used to define the values of filter matrix, stride value, padding value, $h_0$ vector and $c_0$ vector respectively. In case the layer has more than one parameters of the same type, indexes are used to distinguish them. For instance, the two filter matrices in a \emph{ResNet2l} layer can be defined with two keys \emph{filters1} and \emph{filters2}.

For the key \emph{func}, the value must be one of predefined function names. Currently, \name~supports a range of functions which are commonly applied in defining neuron networks, including popular activation functions (e.g., ReLU, Sigmoid, Tanh, and Softmax) and transformation functions (e.g., Reshape and Transpose).\\

% For the key \emph{func}, the value must be one of those predefined function names. \name~supports a range of functions which are commonly applied in defining neuron networks. Each function may require additional parameters, which could be specified with additional keys. The details of functions supported in \name~are shown in Table~\ref{funcs}.

\noindent We acknowledge that, for ordinary users, it may be cumbersome to define (or generate using a program) JSON files in the above format. One remedy is to support automatically translating of models trained using popular frameworks such as Tensorflow or PyTorch. That is, a user simply provides a path to a PyTorch pre-trained model file as the value for the key \emph{path}. \name~then loads the model automatically.

\begin{comment}
\begin{table}[t]
\centering
\begin{tabular}{| l | l | l | l |}
\hline
 Function & Parameters & Keys & Values \\ 
 \hline
ReLU & None & None & None \\
Sigmoid & None & None & None \\
Tanh & None & None & None \\
Softmax & None & None & None \\
Reshape & A new shape for the input & \emph{newshape} & A tuple \\
Transpose & A new axes to be permuted & \emph{axes} & A tuple \\
 \hline
\end{tabular}
\caption{Functions supported by \name}
\vspace{-6mm}
\label{funcs}
\end{table}
\end{comment}

\subsection{Assertion Language} \label{assertion}

For traditional software programs, many assertion languages have been developed, ranging from simple state-based assertions, logic such as Hoare logic~\cite{hoare1969axiomatic}, temporal logic~\cite{pnueli1977temporal} and separation logic~\cite{reynolds2002separation}, to formal specification languages such as the Z language~\cite{spivey1992z} and CSP~\cite{hoare1978communicating}. Existing engines however specify neural network properties in an ad hoc way. Neural networks, as a new programming paradigm, have been applied in a variety of applications. In other words, different neural networks are expected to satisfy different specifications. It is thus important that we have a general language for specifying desirable properties of different neural networks. 

Designing an assertion language is highly non-trivial. We must answer questions such as what is considered a behavior of a neural network and what correctness specification we typically associate with the behaviors. As of now, we take a black-box view of neural networks, i.e., the behavior of a neural network is defined by its input/output relationship. In other words, from a correctness point of view, a neural network can be viewed as a function $M$ such that, given an input feature vector $x$, $M(x)$ is the output vector. It is based on this view that we design our assertion language.
% We remark that with the development of more complicated neural networks, we might have to reason about internal states of neural network models, in which case our assertion language must be refined accordingly.

Formally, a property in \name~is in the general form of $\forall \bar{x}.~\psi_{pre} \Rightarrow \psi_{post}$ where $\psi_{pre}$ is a precondition constraining input samples and $\psi_{post}$ is a postcondition constraining the output and label. Both $\psi_{pre}$ and $\psi_{post}$ are specified using a fragment of first-order logic with built-in functions. The syntax is shown in Figure~\ref{logic}, which is designed to balance expressive and efficiency in terms of analysis tasks. For instance, all free variables in the assertion are universally quantified. Furthermore, each variable $x$ is assumed to be an input of the network, i.e., a feature vector of the specified dimension. Each clause may be a conjunction or disjunction of primitive propositions. Each proposition is an (in)equality, in which 
% or its negation. 
% Moreover, in an (in)equality,
each side is a nested application of predefined functions. %The list of built-in functions which are currently supported in \name are shown in Table~\ref{built-in}.

% Note that there are restrictions such as no existential quantifiers, which limit the expressiveness of the language but also the complexity of the verification problem. For instance, supporting nested quantifiers would complicate the verification algorithm significantly. Yet these restrictions do not prevent interesting properties from being specified. 
In the following, we show how to specify a range of properties which are often relevant to neural network applications.
Note that a constant can be represented by the built-in $id$ function.
% Note that for simplicity, we use well-understood syntactic sugars such as $x_0[0] \neq x_1[0]$ to denote $\neg(x_0[0] = x_1[0])$.
\begin{itemize}
    \item A \emph{reachability} property specifies that if an input satisfies certain constraint (e.g., in certain range), the neural network output must satisfy certain constraint (e.g., in certain range). Such constraints have been supported in existing tools such as Reluplex~\cite{katz2017reluplex}. Specifying such constraint in our language is straightforward. For instance, the following is property 2 from~\cite{katz2017reluplex} written in our language.
    \begin{align*}
       \psi_{pre}: &~x[0] \geq 55947.691 \wedge x[3] \geq 1145 \wedge 
    x[4] \leq 60, \\
    \psi_{post}: &~ L(x) \neq 0 
    \end{align*}  where $x$ is a free variable; $[i]$ is the indexed access function (i.e., $x[3]$ is the 4th element in the feature vector); and $L$ is the labeling function (i.e., it returns the index of the maximum value\footnote{\name~also supports another labelling function which returns the index of the minimum value.} 
    in the output vector according to $x$). Intuitively, the property states that if the input satisfies the precondition, the label must not be 0.
    
    \item \emph{Robustness} is a desirable property for many neural networks. Robustness can be further distinguished into local robustness or global robustness. The former has been the subject of neural network verification and testing in many works~\cite{katz2017reluplex,singh2018fast,singh2019abstract}. It states that any sample $x$ which is similar to a particular existing sample $x_0$ must have the same label of $x_0$. It is specified in our language as follows.
    \begin{align*}
        \psi_{pre}: &~d_i(x,x_0) \leq c, \\
        \psi_{post}: &~L(x) = L(x_0)
    \end{align*}
    where $x_0$ is a constant representing an existing sample; $c$ is a constant; $d_i$ is a predefined function which returns the infinity norm distance between two feature vectors. As shown in Figure~\ref{logic}, users may choose to use $d_0$ or $d_2$ to specify the $0$-norm distance or $2$-norm distance alternatively. Global robustness states that similar samples should have the same label. Note that local robustness is defined based on a particular sample whereas global robustness refers to all samples (including those not in the training set). It can be specified as follows.
    \begin{align*}
        \psi_{pre}: &~d_i(x,y) \leq c, \\
        \psi_{post}: &~L(x) = L(y)
    \end{align*}
    where both $x$ and $y$ are universally quantified variables.
    
    \item \emph{Fairness} is a desirable property for neural network models which may have societal impact. While there are many definitions for fairness~\cite{DBLP:conf/nips/JosephKMR16}, one is called individual discrimination, which can be specified as follows. For simplicity, assume that the first feature of the inputs is the only sensitive feature (such as gender or race). 
    \begin{align*}
        \psi_{pre}: &~x[0] \neq y[0] \wedge x[1]=y[1] \wedge \cdots \wedge x[n]=y[n], \\
        \psi_{post}: &~ L(x) = L(y)  
    \end{align*}
    Intuitively, the above states that for all pair of samples, if the two samples differ only by the sensitive feature, their labels must be the same.
    
    % \item \emph{Miscellaneous} properties other than those defined above can be defined in our language as well. For instance, the following specifies a property which is related to the interpretability of the model. Let $p(x)$ and $q(x)$ be two propositions defined based on a sample that are expressible in our assertion language. 
    % \begin{align*}
    %     \psi_{pre}: &~p(x) = p(y) \wedge q(x) = q(y), \\ 
    %     \psi_{post}: &~L(x) = L(y)
    % \end{align*}
    % Intuitively, the above assertion specifies that if two arbitrary samples are indistinguishable by proposition $p$ and $q$, their labels must be the same. In other words, if the property is verified, we successfully show that this neural network model could be interpreted using a simple model, for instance, in form of a decision tree with the proposition $p$ and $q$. 
\end{itemize}

\begin{figure}[t]
\centering
$$
\begin{array}{lll}
\phi & := & \forall \bar{x}.~\psi_{pre} \Rightarrow \psi_{post} \\
\psi & := & \psi_1 \wedge \psi_2 ~|~ \psi_1 \vee \psi_2 ~|~ \varphi \\
\varphi & := & \bar{f_1}(\bar{x_1}, \bar{c_1}) \bowtie \bar{f_2}(\bar{x_2}, \bar{c_2}) \\
\bowtie & := & > ~|~ \geq ~|~ < ~|~ \leq ~|~ = ~|~ \neq \\
% ~|~ f_1(\bar{x_1}, \bar{c_1}) = f_2(\bar{x_2}, \bar{c_2}) ~|~ \neg \varphi \\
f & := & id~~~~~\mbox{(i.e., the identify function)} \\
& & |~ M~~~\mbox{(i.e., the network model application)} \\
& & |~ L~~~~\mbox{(i.e., the labeling function)} \\
& & |~ N~~~\mbox{(i.e., a linear function)} \\
& & |~ [i]~~~\mbox{(i.e., the indexed access function)} \\
& & |~ d_0~~~\mbox{(i.e., 0-norm distance function)} \\
& & |~ d_2~~~\mbox{(i.e., 2-norm distance function)} \\
& & |~ d_i~~~\mbox{(i.e., infinity norm distance function)} 
\end{array}
$$
\caption{Syntax for assertion language where $\psi_{pre}$ is a precondition; $\psi_{post}$ is a postcondition; $\varphi$ is a primitive proposition; $\bar{x}$ is a set of input variables;
$\bar{c}$ is a set of constants; $\bar{f}$ is a nested application of built-in functions.}
% \vspace{-4mm}
% ; $f$ is a built-in function.}
\label{logic}
\end{figure}

\begin{comment}
\begin{table}[t]
\begin{center}
\begin{tabular}{| l | l |}
\hline
 Function & Definition \\ 
 \hline
 $id$ & The identify function \\
 $M$ & The network model application \\
 $L$ & The labelling function \\
 $N$ & The linear transformation \\
 $[i]$ & The indexed access function \\
 $d_0$ & The 0-norm distance function \\
 $d_2$ & The 2-norm distance function \\
 $d_i$ & The infinity-norm distance function \\
 \hline
\end{tabular}
\caption{The built-in functions}
\label{built-in}
\end{center}
\end{table}
\end{comment}

\noindent \emph{Syntactic Sugars} We acknowledge that it may be difficult for ordinary users to write properties formally. Thus, \name~provides another way to define commonly used properties via predefined templates.
One example template is the robustness property, which can be defined using the following keys.

{\footnotesize \begin{lstlisting}[language=json,firstnumber=1]
  "robustness": "local" | "global",
  "x0": "a floating-point vector",
  "distance": "d0" | "d2" | "di",
  "eps": "a floating-point number"
  "fairness": "an integer vector"
\end{lstlisting}}

With the above keys, users can choose to check either local or global robustness. In case of local robustness, users can specify the value of the original sample $x0$. Users can also choose different distance functions and define the maximum value for the distance. One example of applying the template has been presented in Section~\ref{caseStudy}. %For another example, \name supports syntactic sugar to list indexes of sensitive features in case there are multiple sensitive features.

\subsection{A Benchmark Repository} \label{bench}
To demonstrate that our JSON format is expressive enough to capture real-world analysis tasks, we build a repository of neural network analysis tasks based on the above-described JSON format. Analysis tasks (i.e., a neural network model together with a desired property) which have been reported in various publications have been systematically collected and transformed into the JSON format.
Furthermore, due to the limited type models that previously studied, we additionally train multiple new models, which are added into the repository as well.
% Each task is labeled with the expected verification result.
So far we have not encountered any task which cannot be transformed. 

The goal is also to build a large repository of analysis tasks which can serve as a standard comprehensive benchmark for neural network research community. As of now, the repository contains a set of 12347 tasks, all of which can be downloaded at~\cite{benchurl}. In a nutshell, the repository contains the following network models. 

\begin{itemize}
    \item 45 multilayer perceptron networks and 10 properties used in the experiments of Reluplex~\cite{katz2017reluplex}.
    \item 21 multilayer perceptron and convolutional networks and local robustness property with 100 samples from 2 datasets used in the experiments of DeepZ~\cite{singh2018fast}, DeepPoly~\cite{singh2019abstract},
    RefineZono~\cite{singh2018boosting}, and
    RefinePoly~\cite{singh2019beyond}.
    \item 3 multilayer perceptron networks and fairness property with 100 samples from 3 datasets used in the experiments of ADF~\cite{zhang2020white}.
    \item 4 recurrent networks and local robustness property with 100 samples from 2 datasets, trained by us.
    \item 1 convolutional network and local robustness property with 10000 samples in MNIST challenge~\cite{mnist-challenge}.
\end{itemize}

\section{\name's Analysis Engines} \label{sec:algos}
With the JSON format and the assertion language, a variety of neural network analysis problems can be expressed in \name. Solving them is however the real problem. While we do not claim that \name~is or will be able to solve all of them, we do hope that it provides a platform for researchers to jointly experiment and develop ever-more capable algorithms. As of now, \name~has integrated five analysis engines, such as the verification algorithm known as DeepPoly~\cite{singh2019abstract}, a verification algorithm which extends DeepPoly with abstraction refinement, and the verification algorithm reported in~\cite{tacas2021}. Furthermore, observing that existing approaches are limited to restrictive classes of neural networks, we develop two new engines which are applicable to all neural network models and properties that are currently supported in \name. In the following, we describe the details of these two engines.  

\subsection{Optimization-based Falsification} \label{solver}

The first engine is an optimization-based falsification (a.k.a.~testing) algorithm which is inspired by existing methods on adversarial perturbation~\cite{cw2017Robustness} and fairness testing~\cite{zhang2020white}. 

Given a model $M$ and an arbitrary property $\phi$ in our assertion language, we compile the network model into a function representation internally. The function takes samples as inputs, and produces the output vectors. The definition of the function is built according to the defined layers, layer-by-layer, based on the type of the layer and the provided parameters.
For instance, with a linear layer, we have $y = f(w * x + b)$ where $x$ is the output of the previous layer; $w$ is the weight matrix; $b$ is the bias vector; $f$ is the activation function; and $y$ is the output of the layer.
The result is a function for which, given some particular input, we can easily observe its output as well as the internal computation details, i.e., inputs and outputs of each layer.  

To falsify the property $\phi$ which is in the form of $\forall \bar{x}.~\psi_{pre} \Rightarrow \psi_{post}$, we aim to identify input samples such that $\psi_{pre}$ is satisfied and $\psi_{post}$ is not, i.e., we find an input sample that satisfies $\psi_{pre} \wedge \neg\psi_{post}$. Our idea is to turn this falsification problem into an optimization problem, i.e., we define a loss function based on the formula $\psi_{pre} \wedge \neg \psi_{post}$ and apply guided-search to identify an input sample which satisfies the formula gradually. Intuitively, the loss function is defined to measure how close an input sample is to violate the property and once it is minimized to 0, we successfully falsify the property. Formally, the loss function is defined systematically according to the syntax of the assertion. That is, given any formula $\psi$,  
\[
loss(\psi) = \left\{
    \begin{array}{ll}
        loss(\psi_1) + loss(\psi_2) & \mbox{if $\psi$ is $\psi_1 \wedge \psi_2$} \\
        loss(\psi_1) * loss(\psi_2) & \mbox{if $\psi$ is $\psi_1 \vee \psi_2$} \\
        loss(\varphi) & \mbox{otherwise}
    \end{array}
\right.
\]
where $\varphi$ is $\bar{f_1}(\bar{x_1}, \bar{c_1}) \bowtie \bar{f_2}(\bar{x_2}, \bar{c_2})$ and $loss(\varphi)$ is defined as follows.
\[
loss(\varphi) = \left\{
    \begin{array}{ll}
        max(0, v_2 - v_1 + k) & \mbox{if $\bowtie$ is $>$} \\
        max(0, v_2 - v_1) & \mbox{if $\bowtie$ is $\geq$} \\
        max(0, v_1 - v_2 + k) & \mbox{if $\bowtie$ is $<$} \\
        max(0, v_1 - v_2) & \mbox{if $\bowtie$ is $\leq$} \\
        max(0, |v_1 - v_2|) & \mbox{if $\bowtie$ is $=$} \\
        nid(v_1, v_2, k) & \mbox{if $\bowtie$ is $\neq$} \\
    \end{array}
\right.
\]
where $v_1$ and $v_2$ are values of $\bar{f_1}(\bar{x_1}, \bar{c_1})$ and $\bar{f_2}(\bar{x_2}, \bar{c_2})$ according to the current value of $\bar{x}$, $k$ is a small positive number (e.g., $10^{-9}$) and $nid(a,b,k)$ is 0 if $a \neq b$; otherwise it is $k$. The general idea is the loss function for $\varphi$ should be $0$ if the clause is satisfied and be positive otherwise. The number $k$ guarantees that the value of $loss(\varphi)$ only reaches $0$ when $\varphi$ is satisfied. Moreover, the positive value should show how close the input samples are to satisfy the clause.

The falsification problem thus becomes the following optimization problem. $$\underset{\bar{x}}{\arg\min}~{loss(\psi_{pre} \wedge \neg\psi_{post})}$$
There are many techniques which can be applied to solve this optimization problems. In \name, by default we solve the above constraint optimization problem using the L-BFGS-B algorithm (as there is a mature implementation available in the \emph{scipy} library). 
The algorithm uses the gradient and the estimate of the inverse Hessian matrix of the objective function to guide the search for the minimum value, while maintaining the simple range constraints for variables. The readers are referred to~\cite{byrd1995limited} and~\cite{zhu1997algorithm} for details of the L-BFGS-B algorithm.

% Note that the algorithm terminates when the projected gradient or the change in the value of the objective function is less than a predefined threshold.
% Because of that reason, this falsification engine produces three kinds of results, i.e., successful falsification with a counterexample, successful termination without a counterexample, and timeout. Note that there is a subtle difference between the latter two results. While there is no guarantee that there is no counterexample (i.e., the property is verified) when the algorithm terminates without a counterexample, it arguably provides slightly more `evidence' that the property may be true, compared to the case of a timeout (where optimization is still on-going and thus may find a counterexample if more time is given).

\begin{example}
In the following, we show how the above-described approach works through falsifying a fairness property. The model is a six-layer MLP and is used to predict the income of an adult. The model is trained with the Census Income dataset~\cite{census}. In the dataset, the input has 13 features, which represent personal information of an adult. Among them, 3 features at index 0, 7, and 8 are sensitive features, which represent age, race, and gender respectively. The output is one of the 2 labels, which represents whether the income of an adult is above \$50000. The model can be easily represented in our JSON format. For simplicity, assume that the property to falsify is a local fairness property, i.e., all samples which are different from the given sample $x0 = [4, 0, 7, 0, 0, 4, 2, 0, 1, 5, 0, 40, 0]$ by only the sensitive features have the same label.

\begin{comment}
The property then can be described as:

{\footnotesize \begin{lstlisting}[language=json,firstnumber=1]
{
  "robustness": "local",
  "x0": "[4, 0, 7, 0, 0, 4, 2, 0, 1, 5, 0, 40, 0]",
  "fairness": "[0, 7, 8]"
}
\end{lstlisting}}
\end{comment}

Other than the range constraints for valid inputs (which is defined in the model), the precondition $\psi_{pre}$ is $x[1] = 0 \wedge x[2] = 7 \wedge \cdots \wedge x[6] = 2 \wedge x[9] = 5 \wedge \cdots \wedge x[12] = 0 \wedge (x[0] \neq 4 \lor x[7] \neq 0 \lor x[8] \neq 1)$. The label of the given input is 1 and thus the postcondition is $\psi_{post} : L(x) = 1$. The optimization engine then tries to generate a sample $x$ which satisfies $\psi_{pre}$ but violates $\psi_{post}$. After less than 1 second, a sample $x$ is found with value $x = [2.8769, 0, 7, 0, 0, 4, 2, 0, 1, 5, 0, 40, 0]$ and output vector for $x$ is $[0.5486, 0.4418]$, which results a label of 0. With this result, we can conclude that the model is not locally fair around the given sample. $\hfill$
\end{example}

\subsection{Statistical Model Checking}\label{smc}
The second engine we develop in \name~is based on statistical model checking (SMC~\cite{agha2018survey}). Note that SMC is chosen as it can be applied to all models and properties. Furthermore, it is a formal verification technique that is proven to be effective in combating the complexity of real-world systems, such as cyber-physical systems~\cite{DBLP:conf/atva/ClarkeZ11}.

Given a property $\phi$ in the form of $\forall \bar{x}.~\psi_{pre} \Rightarrow \psi_{post}$, SMC systematically evaluates the probability of those input samples that satisfy $\psi_{pre}$ and violate $\psi_{post}$, through a form of hypothesis testing. To apply SMC, the users are required to provide 4 parameters, which include the expected confidence that the network satisfies the desired properties with probability $\theta$, the bound of indifferent region $\delta$, the values of type I error $\alpha$ and type II error $\beta$. 
% Note that the last three parameters are standard parameters for hypothesis testing and thus default values (e.g., $\delta=0.005$, $\alpha=0.05$ and $\beta=0.05$ according to common practice~\cite{}) are provided. 
With these parameters, hypothesis testing based on the SPRT algorithm (i.e., sequential probability ratio test~\cite{wald1945sequential}) is applied.
% to solve the verification task.
The SPRT algorithm works by generating independent and identically  distributed (IID) random samples to test the following $2$ hypotheses.
\begin{itemize}
    \item $H_0$: The network satisfies $\phi$ with probability $p \geq p_0$ and $p_0 = \theta + \delta$.
    \item $H_1$: The network satisfies $\phi$ with probability $p \leq p_1$ and $p_1 = \theta - \delta$.
\end{itemize}
The details of the SPRT algorithm are shown in Algorithm~\ref{sprt}. Initially, the ratio $pr$ is set to $1$. With the desired property $\phi$, we keep generating IID random samples. If a sample $\bar{x}$ does not satisfy the precondition $\psi_{pre}$, it is discarded and a new sample is generated. In case the sample satisfies the precondition, it is checked against the postcondition $\psi_{post}$. Based on the result, the value of $pr$ is updated. Whenever the ratio value reaches the threshold $\beta / (1 - \alpha)$, $H_0$ is accepted, which means that the probability of the property being satisfied is at least $\theta$ (with a statistical confidence defined by the parameters). Similarly, $H_1$ is accepted whenever the ratio reaches the threshold $(1 - \beta) / \alpha$.

% The above algorithm has one issue. That is, when $\psi_{pre}$ is complicated, it may generate many samples which do not satisfy $\psi_{pre}$, and as a result, take a lot of time. This issue can be partially solved by adopting sampling techniques such as hit-and-run~\cite{DBLP:journals/orl/ChenS96} or QuickSampler~\cite{dutra2018efficient}. The basic idea of these techniques is to apply methods like constraint solving to generate multiple seeds and apply mutation to generate samples based on the seeds. We omit the details as sampling techniques are beyond the content of this paper. %The readers are referred to~\cite{dutra2018efficient} for details.  

\begin{example}
In the following, we present how we apply the above-described SMC to check property 2 with the network ACASXU\_2\_1 reported in~\cite{katz2017reluplex}. The network is an MLP with 7 layers. The input has 5 features and the output has 5 labels. In this example, we assume that the values of the SMC parameters are set as follows: $\theta = 0.95$, $\alpha = 0.05$, $\beta = 0.05$, $\delta = 0.005$. As shown in Section~\ref{assertion}, the property is specified as follows.
\begin{align*}
\psi_{pre}: &~x[0] \geq 55947.691 \wedge x[3] \geq 1145 \wedge 
    x[4] \leq 60, \\
\psi_{post}: &~ L(x) \neq 0
\end{align*}
Note that in addition to $\psi_{pre}$ shown above, all the input features are associated with a range constraint which are omitted for simplicity. Generating IID samples randomly to satisfy $\psi_{pre}$ is straightforward in this example as we simply generate a random value with its range. Applying the SPRT algorithm shown in Algorithm~\ref{sprt}, in less than 1 second, the value of $pr$ reaches $\beta / (1 - \alpha)$ after 300 samples. As a result, the hypothesis $H_0$ (i.e., the network satisfies $\phi$ with probability $p \geq 0.955$) is accepted. 
We remark that accepting $H_0$ does not mean that the property is verified. 
In fact, using the falsification engine introduced in Section~\ref{solver}, we find an adversarial sample $x = [0.6, 0, 0, 0.45, 0.45]$ with output vector $[0.0343, -0.0230, 0.0210, -0.0179, 0.0223]$ which is labeled 0 after less than 1 second. Note that to have better confidence on the correctness of the property, a $\theta$ value arbitrarily closer to 1 can be adopted.
% For instance, with $\theta = 0.99$, the hypothesis $H_1$ is accepted after 222 samples in less than 1 second.
\end{example}

\begin{algorithm}[t]
\SetAlgoLined
% \KwResult{Write here the result }
\KwIn{$\phi = \forall \bar{x} \cdot \psi_{pre} \implies \psi_{post}$}
 $pr = 1$\;
 \While{True}{
  Generate an IID random sample $\bar{x}$\;
  \If{$\bar{x}$ does not satisfy $\psi_{pre}$} {
    continue\;
  }
  \If{$\bar{x}$ satisfies $\psi_{post}$} {
    $pr = pr * p_1 / p_0$\;
  }\Else{
    $pr = pr * (1 - p_1) / (1 - p_0)$\;
  }
  \If{$pr \leq \beta / (1 - \alpha)$} {
    Accept $H_0$\;
  }\ElseIf{$pr \geq (1 - \beta) / \alpha$} { 
    Accept $H_1$\;
  }
 }
 \caption{SPRT algorithm}
 \label{sprt}
\end{algorithm}

\subsection{Experiment Results}
\label{opti-res}
In the following, we systematically evaluate the effectiveness and efficiency of the two engines using benchmark analysis tasks in our repository  presented in Section~\ref{bench}. All the experiments are performed by a machine with 3.1Ghz 8-core CPU and 64GB RAM.    
The results are summarized in Table~\ref{combine1}. Note that ours is the only engine which performs statistical model checking of neural networks to the best of our knowledge. An overall comparison between our engines and all existing ones are made possible with our framework, although it is not the focus of this paper. 

\begin{table*}[t]
\begin{center}
\small
\setlength{\tabcolsep}{2.6pt}
\begin{tabular}{| l | l | c | c | c | c | c | c | c | c | c | c | c | c | c |}
\hline
 \multirow{2}{*}{\textbf{Prop.}} & \multirow{2}{*}{\textbf{Networks}} & \multirow{2}{*}{\textbf{\#Tasks}} & \multicolumn{3}{c|}{\textbf{Optimization}} & \multicolumn{3}{c|}{\pmb{SMC $\theta = 0.90$}} &
 \multicolumn{3}{c|}{\pmb{SMC $\theta = 0.95$}} &
 \multicolumn{3}{c|}{\pmb{SMC $\theta = 0.99$}} \\
 \cline{4-15}
 & & & \textbf{V$^*$} & \textbf{F} & \textbf{Time} & \textbf{H0} & \textbf{H1} & \textbf{Time} & \textbf{H0} & \textbf{H1} & \textbf{Time} & \textbf{H0} & \textbf{H1} & \textbf{Time} \\
 \hline
 P1~\cite{katz2017reluplex}  & ACASXU\_*\_* (all networks) & 45 & 45 & 0 & 32s & 45 & 0 & 30s & 45 & 0 & 30s & 45 & 0 & 30s \\
 P2~\cite{katz2017reluplex}  & ACASXU\_$x$\_* $(x \geq 2)$ & 36 & 8 & 28 & 23s & 36 & 0 & 24s & 36 & 0 & 24s & 14 & 22 & 24s \\
 P3~\cite{katz2017reluplex}  & ACASXU\_*\_* $\neq (1\_\{7,8,9\})$ & 42 & 42 & 0 & 29s & 42 & 0 & 28s & 42 & 0 & 28s & 42 & 0 & 28s \\
 P4~\cite{katz2017reluplex}  & ACASXU\_*\_* $\neq (1\_\{7,8,9\})$ & 42 & 42 & 0 & 29s & 42 & 0 & 28s & 42 & 0 & 28s & 42 & 0 & 28s \\
 P5~\cite{katz2017reluplex}  & ACASXU\_1\_1 & 1  & 1 & 0 & 1s & 1 & 0 & 1s & 1 & 0 & 1s & 1 & 0 & 1s \\
 P6~\cite{katz2017reluplex}  & ACASXU\_1\_1 & 1  & 1 & 0 & 1s & 1 & 0 & 1s & 1 & 0 & 1s & 1 & 0 & 1s \\
 P7~\cite{katz2017reluplex}  & ACASXU\_1\_9 & 1  & 1 & 0 & 1s & 1 & 0 & 1s & 1 & 0 & 1s & 1 & 0 & 1s \\
 P8~\cite{katz2017reluplex}  & ACASXU\_2\_9 & 1  & 1 & 0 & 1s & 1 & 0 & 1s & 1 & 0 & 1s & 1 & 0 & 1s \\
 P9~\cite{katz2017reluplex}  & ACASXU\_3\_3 & 1  & 1 & 0 & 1s & 1 & 0 & 1s & 1 & 0 & 1s & 1 & 0 & 1s \\
 P10~\cite{katz2017reluplex} & ACASXU\_4\_5 & 1  & 1 & 0 & 1s & 1 & 0 & 1s & 1 & 0 & 1s & 1 & 0 & 1s \\
 Robust. & MNIST\_ReLU\_4\_1024 & 98 & 23 & 75 & 33s & 96 & 2 & 26s & 96 & 2 & 26s & 95 & 3 & 27s \\
 Robust. & MNIST\_ReLU\_6\_100 & 99  & 10 & 89 & 17s & 99 & 0 & 7s & 99 & 0 & 7s & 99 & 0 & 7s \\
 Robust. & MNIST\_ReLU\_9\_200 & 97 & 10 & 87 & 15s & 93 & 4 & 10s & 92 & 5 & 10s & 90 & 7 & 9s \\
 Robust. & MNIST\_Sigmoid\_6\_500 & 95 & 4 & 91 & 13s & 93 & 2 & 17s & 93 & 2 & 18s & 92 & 3 & 19s \\
 Robust. & MNIST\_Sigmoid\_6\_500\_PGD\_0.1 & 100 & 90 & 10 & 3m47s & 98 & 2 & 18s & 98 & 2 & 19s & 98 & 2 & 19s \\
 Robust. & MNIST\_Sigmoid\_6\_500\_PGD\_0.3 & 97 & 84 & 13 & 2m37s & 96 & 1 & 18s & 96 & 1 & 19s & 96 & 1 & 19s \\
 Robust. & MNIST\_Tanh\_6\_500 & 99 & 17 & 82 & 1m4s & 98 & 1 & 17s & 98 & 1 & 18s & 97 & 2 & 18s \\
 Robust. & MNIST\_Tanh\_6\_500\_PGD\_0.1 & 100 & 97 & 3 & 3m8s & 100 & 0 & 18s & 100 & 0 & 18s & 100 & 0 & 19s \\
 Robust. & MNIST\_Tanh\_6\_500\_PGD\_0.3 & 100 & 97 & 3 & 3m5s & 99 & 1 & 19s & 99 & 1 & 18s & 99 & 1 & 19s \\
 Robust. & MNIST\_Conv\_Small\_ReLU & 100 & 89 & 11 & 49s & 100 & 0 & 28s & 99 & 1 & 29s & 99 & 1 & 29s \\
 Robust. & MNIST\_Conv\_Small\_ReLU\_DAI & 99 & 96 & 3 & 1m17s & 99 & 0 & 27s & 99 & 0 & 28s & 99 & 0 & 30s \\
 Robust. & MNIST\_Conv\_Small\_ReLU\_PGD & 100 & 98 & 2 & 55s & 100 & 0 & 27s & 100 & 0 & 29s & 100 & 0 & 29s \\
 Robust. & MNIST\_Conv\_Big\_ReLU\_DAI & 95 & 94 & 1 & 6m35s & 95 & 0 & 2m33s & 95 & 0 & 2m41s & 95 & 0 & 2m47s \\
 Robust. & MNIST\_Conv\_Super\_ReLU\_DAI & 99 & 97 & 2 & 20m24s & 99 & 0 & 8m30s & 98 & 1 & 8m55s & 98 & 1 & 9m3s \\
 Robust. & CIFAR\_ReLU\_6\_100 & 16 & 0 & 16 & 6s & 7 & 9 & 6s & 7 & 9 & 6s & 7 & 9 & 6s \\
 Robust. & CIFAR\_ReLU\_7\_1024 & 16 & 0 & 16 & 34s & 10 & 6 & 43s & 10 & 6 & 41s & 8 & 8 & 39s \\
 Robust. & CIFAR\_ReLU\_9\_200 & 9 & 0 & 9 & 8s & 7 & 2 & 9s & 5 & 4 & 8s & 5 & 4 & 8s \\
 Robust. & CIFAR\_Conv\_Small\_ReLU & 59 & 0 & 59 & 8s & 52 & 7 & 27s & 49 & 10 & 30s & 45 & 14 & 24s \\
 Robust. & CIFAR\_Conv\_Small\_ReLU\_DAI & 53 & 2 & 51 & 12s & 50 & 3 & 31s & 49 & 4 & 24s & 47 & 6 & 24s \\
 Robust. & CIFAR\_Conv\_Small\_ReLU\_PGD & 70 & 0 & 70 & 9s & 64 & 6 & 38s & 64 & 6 & 30s & 63 & 7 & 29s \\
 Robust. & CIFAR\_Conv\_Big\_ReLU\_DAI & 60 & 0 & 60 & 25s & 53 & 7 & 2m53s & 52 & 8 & 2m47s & 48 & 12 & 2m9s \\
 Robust. & Jigsaw\_GRU & 94 & 65 & 29 & 7m1s & 91 & 3 & 4m & 90 & 4 & 4m20s & 88 & 6 & 5m15s \\
 Robust. & Jigsaw\_LSTM & 93 & 69 & 24 & 12m38s & 93 & 0 & 4m25s & 93 & 0 & 4m43s & 92 & 1 & 5m18s \\
 Robust. & Wiki\_GRU & 96 & 55 & 41 & 11m22s & 96 & 0 & 4m49s & 96 & 0 & 4m54s & 96 & 0 & 5m10s \\
 Robust. & Wiki\_LSTM & 94 & 44 & 50 & 29m42s & 93 & 1 & 5m27s & 92 & 2 & 5m37s & 92 & 2 & 5m40s \\
 Fairness & Bank\_MLP\_6\_Layers & 99 & 90 & 9 & 4s & 91 & 8 & 2s & 89 & 10 & 3s & 87 & 12 & 2s \\
 Fairness & Census\_MLP\_6\_Layers & 86 & 62 & 24 & 4s & 61 & 25 & 2s & 56 & 30 & 3s & 45 & 41 & 2s \\
 Fairness & Credit\_MLP\_6\_Layers & 100 & 56 & 44 & 3s & 64 & 36 & 3s & 56 & 44 & 2s & 51 & 49 & 2s \\
\hline
 \textbf{Total} & \textbf{73 networks} & \textbf{2494} & \textbf{1492} & \textbf{1002} & \textbf{1h49m} & \textbf{2368} & \textbf{126} &\textbf{41m6s} & \textbf{2341} & \textbf{153} & \textbf{42m19s} & \textbf{2280} & \textbf{214} & \textbf{43m38s} \\
 \hline
\end{tabular}
\caption{Experiment with reachability, local robustness and local fairness property}
\vspace{-4mm}
% Timeout: 1 hour for each network.}
\label{combine1}
\end{center}
\end{table*}

In the table, the first 2 columns show the properties and the networks under analyze respectively. For the local robustness property, we use infinity norm distance with the maximum value is $0.1$. Note that the model name is coded with information such as the type of network, the activation function and additional parameters. Given a model and a property, there may be many tasks (i.e., different local robustness property for different input samples). The third column shows the number of tasks in each setting. Note that for local robustness and fairness properties, we randomly choose 100 samples from the datasets and for each network only the samples which are classified correctly (by comparing with the provided ground truth) are used in the tasks.

The next column shows the results of \name~optimization engine. The sub-column F shows the number of falsified tasks with counterexamples. Note that \name~may terminate without a counterexample, which does not guarantee that the task is verified (due to the nature of optimization algorithms) and we indicate such results using a sub-column titled V$^*$. Due to the space limit, we are unable to report the results of all 12347 tasks. Rather, a set of 2494 tasks are selected with the objective to cover a variety of networks and properties. From the table, we see that \name~can handle all the 2494 tasks. We remark that \name~is sound when it reports that the properties is falsified. That is, 992 tasks are successfully falsified, with a counterexample. For the remaining tasks, the model is more likely to satisfy the property.

\begin{comment}
Finally, we apply the falsification engine in \name~to generate counterexample for 10000 samples in MNIST challenge, in which 9853 samples are classified correctly. The results show that \name~can generate 473 counterexamples satisfy the requirement of the challenge (i.e., the maximum infinity-norm distance to the original sample is $0.3$) after running 15h4m.
%We notice that the result is much more worse than the other tools reported in the challenge site. However, the result is expected considering that the other tools are specific attacking tools for the problem. Moreover, with the design of \name, we can easily integrate the algorithms of other tools if necessary. 
\end{comment}

The next 3 columns show the results of statistical model checking engine. Note that this engine is designed for probabilistic verification rather than falsification.
% Again, we apply the engine to all the verification tasks discussed in Section~\ref{bench}. To the best of our knowledge, \name~is the only statistical model checker for neuron networks and thus we have no baseline to compare with. 
As the SPRT algorithm is parameterized with the probability $\theta$, we apply it to all the tasks with 3 different values of $\theta$ (i.e., $0.90$, $0.95$, and $0.99$) so that we can observe the effort required to reach different level of statistical confidence. The same `default' values are adopted for the remaining 3 parameters, i.e., $\alpha = 0.05$, $\beta = 0.05$, and $\delta = 0.005$.

In the last 3 columns, for each value of $\theta$, the sub-columns H0 and H1 show the number of tasks in which the hypothesis $H_0$ and $H_1$ are accepted respectively. As we can see, when the value of $\theta$ increases, the number of tasks in which $H_0$ is accepted decreases while the number of tasks in which $H_1$ is accepted increases. This is an expected result considering that the users want to be more confident about the verified properties with a higher $\theta$ value.
% We also notice that the numbers of generated samples for $\theta = 0.90$, $0.95$, and $0.99$ are 709874, 719668, and 695373 respectively. We see that there is a reluctant between these numbers. The result is reasonable considering that when $\theta$ increases, we need more samples to accept $H_0$ but less samples to accept $H_1$. Moreover, some specific tasks may make the value of $pr$ go back and forth (i.e., need more samples) before reaching the threshold. The result helps to confirm that the number of generated samples depends on the specific task rather than the value of $\theta$ alone.
Finally, for the time needed to run the experiment, we do not see any significant difference between the 3 settings of the statistical model checking engine. All of them can finish testing 2494 tasks in less than 1 hour. However, we notice that the running time of the statistical model checking engine is less than half of the time needed by the optimization engine.

\begin{comment}
For the last experiment, we apply the testing engine in \name~to test the local robustness of 10000 sample in the MNIST challenge. With the setting $\theta = 0.90$, the engine accepts $H_0$ 9760 times and accept $H_1$ 93 times. With $\theta = 0.95$, $H_0$ is accepted 9724 times and $H_1$ 129 times. These numbers are 9657 and 196 respectively for $\theta = 0.99$. Again, we can see that the number of times $H_0$ is accepted decreases when the value of $\theta$ increases. In the experiment, the engine generates more than 2.6 million samples for $\theta = 0.90$, and more than 2.8 million samples for $\theta = 0.95$ and $0.99$. The time needed to run the experiment is 4h38m, 4h51m, and 4h54m for $\theta = 0.90$, $0.95$, and $0.99$ respectively.
% , which is again much less than the time needed by the falsification engine.
\end{comment}

\section{Related Work}
\label{sec:related}
% In this section, we briefly review existing approaches on verification and falsification of neural networks and discuss how our approach is different. 

This work is closely related to existing approaches on verifying local robustness of neural networks. 
% Different from challenging the neural network from destructive perspectives i.e. adversary attacks, 
Ever since the discovery of adversarial samples~\cite{fgsm}, verifying robustness of neural networks attracted much attentions due to their implications in safety critical applications. Existing approaches can be roughly classified into two groups: exact methods and approximation methods. 

The exact methods aim to capture semantics of neural networks precisely and solve the verification problem through constraint solving. In~\cite{tjeng2017evaluating}, Tjeng \emph{et al.} proposed to tackle the problem using Mixed Integer Linear Programming (MILP). In~\cite{katz2017reluplex,ehlers2017formal}, the authors proposed to solve the problem through SMT solving. These methods can verify a neural network as long as the property holds. The limitations of these methods are that they are limited to analyze
% (feedforward) neural 
networks with ReLU activation functions only. In other words, popular activation functions such as Sigmoid and Tanh are not supported. Furthermore, these methods typically have limited scalability.
% , i.e., they can only handle networks with a small number of layers and neurons, as we partly demonstrate in Section~\ref{opti-res}.

The approximation methods leverage well-developed techniques such as linear approximations~\cite{weng2018towards,wong2017provable} and abstract interpretation~\cite{gehr2018ai2,singh2018fast,singh2019abstract}. The idea is to conduct an over-approximation of the given neural network (e.g., through over-approximating each neuron with a simple linear constraint), and verify properties soundly based on the over-approximation. Thanks to the linear approximation, these approaches (although not all of them) could handle a wider range of activation function such as ReLU, Sigmoid or Tanh. Furthermore, these approaches are typically more scalable than the exact methods. However, due to the over-approximation, these methods are sound but not complete, i.e., they may fail to verify a valid property due to the presence of the spurious counterexamples.

% Our project does not aim to replace these impressive efforts. Rather, we aim to provide a platform which integrates and further develops these efforts. Furthermore, the two new verification engines supported in \name~are complementary to existing verification approaches, i.e., the falsification engine would allow us to efficiently falsify those properties which are not satisfied whereas the SMC engine provides a way of verifying neural network probabilistically.  

Beside robustness, this work is related to a line of work on analyzing fairness of neural networks, collectively called fairness testing.
Several approaches have been proposed on fairness testing machine learning models including neural networks. All of them search for discriminatory instances (i.e., counterexamples to fairness) through certain heuristic-based sampling techniques. Galhotra \emph{et al.} proposed THEMIS~\cite{angell2018themis, galhotra2017fairness}, a causality based algorithm utilizing the random test generation to evaluate a model's fairness, i.e., the frequency of individual discriminatory instances.
% Their work could be viewed as a heuristic-based approach for statistical model checking of fairness.
Udeshi \emph{et al.} proposed AEQUITAS~\cite{udeshi2018automated} which is based on THEMIS. AEQUITAS works in two phases, i.e., a `global search' phase, which attempts to explore the whole input domain, followed by a `local search' phase, which searches within the neighboring region of the instances identified in the global phase. Zhang \emph{et al.} proposes a lightweight algorithm ADF to efficiently generate individual discriminatory instances~\cite{zhang2020white}. They perturb instances near the decision boundary in the global search and leverage the gradient information to guide the local search.

% We note that existing approaches focuses on testing of fairness rather than fairness verification. To the best of our knowledge, ours is the first attempt to support fairness verification on neural networks. 

The falsification engine in \name~is inspired by the many adversarial sample generation methods. Since Szegedy \emph{et al.} discovered that neural networks are vulnerable to adversarial samples~\cite{fgsm}, many attacking methods have been developed to generate adversarial samples efficiently with minimal perturbation. Some examples are the FGSM method~\cite{fgsm}, the Jacobian-based saliency map attack~\cite{DBLP:conf/eurosp/PapernotMJFCS16}, and C\&W~\cite{cw2017Robustness}. Lastly, this work is remotely related to recent papers which proposed different coverage criteria for evaluating the effectiveness of a test set, along with different methods to generate test cases to improve the coverage criteria. For instance, DeepXplore~\cite{deepXplore} proposed the first testing criterion for DNN models, i.e., Neuron Coverage (NC), which calculates the percentage of activated neurons (w.r.t. an activation function) among all neurons.
% Unlike the above-mentioned work, this project focuses on verification of neural networks. 

\section{Conclusion} \label{sec:conclude}
In this work, we aim to develop a unified platform for neural network analysis. Towards our goal, we make three technical contributions. First, we propose a unified JSON format for capturing a variety of neural network models as well as an assertion language for specifying neural network properties. We further build a repository of 12347 tasks, which serves as a comprehensive benchmark for evaluating neural network analysis techniques. Second, we develop two novel algorithms for tackling the falsification/verification problems for a variety of models and properties. The experiment results show that these algorithms can handle a wide range of networks and properties. Lastly, we devote non-trivial amount of engineering effort to make our project a useful open source platform. 

We are continuously developing our platform further as the following activities: integrating more analysis algorithms, extending them with further optimizations (such as abstraction refinement for existing abstraction interpretation based approaches) and developing new algorithms (such as probabilistic model checking for neural networks).   

\bibliographystyle{IEEEtran}
\bibliography{IEEEabrv,refs}
% that's all folks
\end{document}